\documentclass[11pt,a4paper]{article}
\usepackage[hyperref]{acl2020}
\usepackage{times}
\usepackage{latexsym}
\usepackage{graphicx}

\usepackage{microtype}
\usepackage[normalem]{ulem} 
\usepackage{hyperref}

\aclfinalcopy 
\def\aclpaperid{2759} 

\title{Spying on your neighbors: Fine-grained probing of contextual embeddings for information about surrounding words}

\author{Josef Klafka \\
  Department of Psychology \\
  Carnegie Mellon University \\
  \texttt{jklafka@andrew.cmu.edu} \\\And
  Allyson Ettinger \\
  Department of Linguistics \\
  University of Chicago \\
  \texttt{aettinger@uchicago.edu} \\}

\date{}

\begin{document}
\maketitle
\begin{abstract}
Although models using contextual word embeddings have achieved state-of-the-art results on a host of NLP tasks, little is known about exactly what information these embeddings encode about the context words that they are understood to reflect. To address this question, we introduce a suite of probing tasks that enable fine-grained testing of contextual embeddings for encoding of information about surrounding words. We apply these tasks to examine the popular BERT, ELMo and GPT contextual encoders, and find that each of our tested information types is indeed encoded as contextual information across tokens, often with near-perfect recoverability---but the encoders vary in which features they distribute to which tokens, how nuanced their distributions are, and how robust the encoding of each feature is to distance. We discuss implications of these results for how different types of models break down and prioritize word-level context information when constructing token embeddings. 
\end{abstract}

\section{Introduction}

The field of natural language processing has recently seen impressive performance gains associated with the use of ``contextual word embeddings'': high-dimensional vectors that have access to information from the contexts of the words they represent. Models that use these contextual embeddings achieve state-of-the-art performance on a variety of natural language processing tasks, from question-answering to natural language inference. As of writing, nearly all of the models on the SuperGLUE leaderboard~\cite{wang2019superglue} use contextual embeddings in their architectures, most notably models building on the BERT~\cite{devlin2018bert} and Transformer XL~\cite{dai2019transformer} models. 

Despite the clear power afforded by incorporating context into word embeddings, little is known about what information these contextual embeddings actually encode about the words around them. In a sentence like ``The lawyer questioned the judge'', does the contextual representation for \emph{questioned} reflect properties of the subject \emph{lawyer}? Of the object \emph{judge}? What determines the information that a contextual embedding absorbs about its surrounding words? In this paper, we address these questions by designing and implementing a suite of probing tasks, to test contextual embeddings for information about syntactic and semantic features of words in their contexts. We use controlled sentences of fixed structure, allowing us to probe for information associated with word categories, and to avoid confounds with particular vocabulary items. We then apply these tests to examine the distribution of contextual information across token representations produced by contextual encoders BERT~\cite{devlin2018bert}, ELMo~\cite{peters2018deep}, and GPT~\cite{radford2018improving}.

The contributions of this paper are twofold. First, we introduce a suite of novel probing tasks for testing how encoders distribute contextual information across sentence tokens. All datasets and code are available for follow-up testing.\footnote{Probing datasets and code available at \hyperlink{https://github.com/jklafka/context-probes}{https://github.com/jklafka/context-probes}.} 
Second, we use these tests to shed light on the distribution of context information in state-of-the-art encoders BERT, ELMo and GPT. We find that these models encode each of our tested word features richly across sentence tokens, often with perfect or near-perfect recoverability, but the details of how the models distribute this information vary across encoders. In particular, bidirectional models show more nuance in information selectivity, while the deeper transformer models show more robustness to distance. Follow-up tests suggest that the effects cannot be chalked up to proximity, and that general word features are encoded more robustly than word identity.

\section{Our approach}

Our tests address the following basic question: if we probe the contextual representation of a given token in a sentence, how much information can we recover about the other words in that sentence? For example, if we create a contextual embedding for the word \emph{questioned} in the sentence
\begin{center}
    \emph{The lawyer questioned the judge}
\end{center}
how well can we extract information about the subject noun (\emph{lawyer})? What if we probe the object noun (\emph{judge}) or determiners (\emph{the})? We develop tasks to probe representations of each word for various types of information about the other words of the sentence, allowing us to examine with fine granularity how contextual encoders distribute information about surrounding words. We complete this investigation for each word in a set of fixed-length sentences of pre-determined form, which allows us to characterize behaviors based on word categories (e.g., subjects versus verbs). Using this approach, we can examine how the distribution of context information is impacted by a) the type of information being encoded, and b) the properties of the word that the embedding corresponds to. 

\begin{table*}[t!]
 \begin{tabular}{c c c} 
 Task & Example & Label \\
 \hline\hline
 (subject) Number & \emph{The \textbf{lawyer} betrayed the judge.} & SINGULAR \\ 
 & \emph{The \textbf{lawyers} betrayed the judge.} & PLURAL \\
 \hline
 (subject) Gender & \emph{The \textbf{waiter} betrayed the judge.} & MASCULINE \\
  & \emph{The \textbf{waitress} betrayed the judge.} & FEMININE \\
 \hline
 (subject) Animacy & \emph{The \textbf{car} betrayed the judge.} & INANIMATE \\
  & \emph{The \textbf{turtle} betrayed the judge.} & ANIMATE \\
 \hline
\end{tabular}
\centering
\caption{Example items from probing tasks for each \textbf{noun} information type.}
\label{tab:nouns}
\end{table*}

\section{Related work}
Much work has been done on analyzing the information captured by sentence encoders and language models in general. Classification-based probing tasks have been used to analyze the contents of sentence embeddings~\cite{adi2016,conneau2018you,ettinger2016probing}, finding that these embeddings encode a variety of information about sentence structure, content, length, etc., though more tightly-controlled tasks suggest weaknesses in capturing basic sentence meaning~\cite{ettinger2018assessing}. Our work uses the same classification-based probing methodology, but focuses on probing token-level embeddings for context information.

Other work has analyzed linguistic capacities of language models by examining output probabilities in context, emulating methods for studying human language processing. Much of this work has studied sensitivity to syntactic dependencies in recurrent neural network language models~\cite[e.g.][]{linzen2016assessing, wilcox2018rnn,chowdhury2018rnn,gulordava2019colorless, marvin2019targeted,futrell2019neural}. Using similar methods to test syntactic awareness in BERT,~\citet{goldberg2019assessing} finds the model to perform almost at ceiling on syntactic tests. Testing BERT's outputs on a range of semantic, syntactic and pragmatic information,~\citet{ettinger2020bert} finds strong sensitivity to syntax, but clear limitations in areas of semantics and pragmatic/commonsense reasoning. We complement this work with a direct focus on the contextual token representations learned by models pre-trained on language modeling, examining the syntactic and semantic information that these embeddings capture about surrounding words. 

Most directly related to the present work are studies using probing and other methods to analyze information in contextual token embeddings. Some of this research \cite[e.g.][]{tenney2019bert, jawahar2019does} finds that BERT encodes more local, syntactic information at lower layers and more global, semantic information at higher layers. \citet{peters2018dissecting} find that encoders differ in encoding strength for semantic features but all encode these features strongly where possible. \citet{hewitt2019structural} provide evidence that contextual encoders capture sentence-level hierarchical syntactic structures in their representations. Other work \cite{liu2019linguistic, tenney2019you} finds that contextual word encoders struggle to learn fine-grained linguistic information in a variety of contexts. These papers have focused primarily on studying the ability of contextual embeddings to capture information about the full sentence, or about phrases or dependencies of which those contextual embeddings are a part. We focus on mapping the precise distribution of context information across token embeddings, with a systematic, fine-grained investigation of the information that each token encodes about each of its surrounding tokens.


\section{Probing for contextual information}

For each of our probing tasks, we test for a particular \emph{information type}, formulated as a query about a particular \emph{target word} in the sentence---for instance, ``What is the animacy of the subject?'' or ``What is the tense of the verb?''. We then apply these queries to probe the embeddings for each word of the sentence in turn---we call this the \emph{probed word}. For example, a test with a \emph{probed word} of verb, a \emph{target word} of subject noun, and an \emph{information type} of animacy would ask the question: ``What does the embedding of the \emph{verb} tell us about the \emph{animacy} of the \emph{subject noun}?'' We implement each test as a binary classification task (e.g., ``animate'' vs ``inanimate''), and train and test a multi-layer perceptron classifier using the embeddings of one probed word category at a time as input for the task. In this section, we describe the details of our probing datasets and tested information types. 

\subsection{Dataset construction}

We construct our datasets using generated transitive sentences with a fixed five-word structure: ``DET SUBJ-N VB DET OBJ-N'', as in ``\emph{The lawyer questioned the judge}''. For generating these sentences, we draw nouns and verbs from the intersection of the single-word vocabularies of the four tested encoding models, from which we select a set of $100$ target words for each task, along with a set of $100$ of each other content word type. We select based on the necessary properties for the individual probing tasks (for example, as shown in Table~\ref{tab:nouns}, the gender task requires explicitly gendered nouns, and the animacy task requires a balanced set of animate vs inanimate nouns). We constrain our sample to ensure balance between positive and negative labels in training and test sets. The stimuli for each task were checked by the first author, a native English speaker, to confirm plausibility of occurrence in a corpus of English text. The exception to the plausibility rule was the noun animacy task, which required certain implausible noun-verb pairings.

We follow~\citet{ettinger2018assessing} in employing controls to keep selected baselines at chance performance---in our case, we ensure that non-contextualized GloVe embeddings \cite{pennington2014glove} are at chance on all tests, except when the probed word is the target word (e.g., when testing ``what does the verb embedding tell us about the verb''). This ensures that the tasks must be solved by incorporating contextual information, rather than by spurious cues in the words themselves. Controlling in this way requires attention to inflectional marking. When targeting subject number we use only past tense transitive verbs (which have the same form regardless of subject number) to ensure that no word but the target noun indicates the number information of interest. 

For each task we generate $4000$ training and $1000$ test transitive sentences. We generate separate datasets for each target word within an information type---for example, generating separate subject animacy and object animacy datasets.

\subsection{Information types}

We probe for three types of linguistic information about nouns and three types of linguistic information about verbs. We select these as reasonably simple and fundamental syntactic and semantic features at the word level, which are thus good candidates to be encoded in representations for other words in the sentence. With our selections, we aim for diversity in how syntactic or semantic the information is, and in whether the targeted information is overtly marked on the target word itself. 

\begin{table*}[t!]
\begin{tabular}{c c c} 
 Task & Example & Label \\
 \hline\hline
 Tense & \emph{The lawyer \textbf{betrayed} the judge.} & PAST \\
  & \emph{The lawyer \textbf{betrays} the judge.} & PRESENT \\
 \hline
 Causative-inchoative & \emph{The warden \textbf{melted} the ice. (the ice melted)} & YES CAUSATIVE/INCHOATIVE \\
 alternation & \emph{The warden \textbf{bought} the ice. (*the ice bought)} & NO CAUSATIVE/INCHOATIVE \\
 \hline
 Dynamic-stative & \emph{The lawyer \textbf{found} the judge.} & DYNAMIC VERB \\
  & \emph{The lawyer \textbf{observed} the judge.} & STATIVE VERB \\
 \hline
\end{tabular}
\centering
\caption{Example items from probing tasks for each \textbf{verb} information type.}
\label{tab:verbs}
\end{table*}

\paragraph{Noun information} When probing for information about subject and object nouns, we target three types of information: \textbf{number}, \textbf{gender}, and \textbf{animacy}. 
The number of a noun in English (whether it is singular or plural) is a basic property that has syntactic implications for verb agreement, and that is directly encoded on the surface form of the noun. 
Gender is a primarily semantic feature, and English nouns sometimes indicate gender in their surface forms (e.g. \emph{actor} versus \emph{actress}), but in other cases they do not (e.g. \emph{brother} versus \emph{sister}). Recent work has examined gender bias in word embeddings \cite[e.g.,][]{caliskan2017semantics}, further highlighting the importance of understanding how this information is reflected in word representations.
Animacy is a semantic property that distinguishes animate entities like humans from inanimate entities like cars, and impacts contextual factors like the kind of verb frames a noun is likely to occur in. 

Table \ref{tab:nouns} shows example items from probing tasks for each of these noun information types---in this case with the subject noun as the target word. 
The first line for each task shows an example of a positive label sentence, and the second line shows an example of a negative label sentence.
We also design probing tasks that target information about the object noun. These tasks are nearly identical in form to the subject tasks: the target word is simply switched to the object, such that the positive and negative labels are determined by the properties of the object noun rather than the subject noun. 

\paragraph{Verb information} When probing for information about verbs, we target three types of information: \textbf{tense}, presence of a \textbf{causative-inchoative alternation}, and classification of \textbf{dynamic versus stative} verbs. 
Tense information in English is a largely semantic property with some syntactic implications, and it is marked by morphology on the surface form of a verb. In our probing tasks, we restrict to testing present versus past tense. In our verb tense task, we only use singular subjects, to avoid information about the subject influencing variation in the verb form. Present verbs encoding subject number is the only situation in which information about one word is explicitly marked on another word in our tasks. For all other tasks, we use only past tense verbs, which don't have surface marking of subject information.
The causative-inchoative alternation refers to whether a verb has both a transitive and an intransitive meaning---this is a syntactic/semantic feature that has essential implications for the way that a verb can interact with its context.\footnote{This task is derived from the verb alternation probe of the same name in \citet{warstadt2018neural}.} 
The dynamic-stative feature is a primarily semantic feature referring to whether a verb involves the subject producing a change in the object (dynamic), or communicates a state of the subject and the object (stative). The causative-inchoative and dynamic-stative feature information are not marked on the surface forms of the verb. 

We have included examples for tasks testing each of these verb information types in Table \ref{tab:verbs}. 

\paragraph{Determiner information}

While we do probe for information \emph{encoded on} our determiner words (\emph{the}), we do not design tests that treat these determiners as \emph{target words}. English determiners are a small closed-class set, making it difficult to design datasets with sufficient variety for probing. We leave this problem for future work.

\section{Experiments}

We apply our probing tasks to test for the distribution of contextual information across tokens in three prominent contextual encoders: BERT\textsubscript{BASE}~\cite{devlin2018bert}, ELMo~\cite{peters2018deep}, and GPT~\cite{radford2018improving}.

BERT\textsubscript{BASE} is a bidirectional transformer architecture of 12 layers, trained on a novel masked language modeling task of predicting randomly masked tokens using left and right context, as well as a next-sentence prediction task. We probe representations from the model's final layer, based on results suggesting that BERT's later layers contain more semantic and abstract information~\cite[e.g.][]{jawahar2019does}. ELMo is composed of stacked bidirectional LSTMs, trained by jointly optimizing backwards and forwards language modeling objectives. We use the original version of ELMo with two representation layers, and we probe representations from the second layer, which has also been found to encode more abstract and semantic information \cite{peters2018deep}. GPT is a unidirectional left-to-right 12-layer transformer, also trained on language modeling. Consistent with ELMo and BERT, we probe representations from GPT's final layer.\footnote{We also test the second-to-last layers from each model, and find that the results differ in magnitude from results on the final layer, but show the same overall patterns.}
We test the pre-trained versions of these models without fine-tuning, to examine their general-purpose encoding capacities, in line with~\citet{peters2018dissecting}. 

We use these models to embed each of our five-word sentences, producing contextualized representations for each token. Then for each probing task (e.g., subject animacy, verb tense) we train and test classifiers on the embeddings for a single probed word category (e.g., object noun) at a time. 

We use several classifier architectures in our probing tasks, in order to explore the impact of classifier complexity on extraction of our target information types. We use a multilayer perceptron classifier with a single hidden layer of $1024$ units, as well as a smaller classifier with three layers of $20$ units each, and a larger classifier with three layers of $1024$ units each. We use the relevant contextual or non-contextual token representations as input for classification. The largest inputs we supply to the classifiers are contextual embeddings with dimension $1024$, from ELMo.We use the relevant contextual or non-contextual token representations as input to the classifiers. Finding similar results across classifier architectures, we follow precedent in the literature~\citep{adi2016,ettinger2018assessing} and present results only for our classifier with a single hidden layer. To quantify variance across runs, we repeat this process $50$ times for each probed word category on each task.\footnote{Training intermittently produced outlier runs with chance-level or below-chance test accuracy in settings with otherwise strong performance---we omit such runs from consideration.}

As a sanity-check baseline, we also test non-contextual GloVe embeddings \cite{pennington2014glove} on each of our tasks, to establish how well each information type is captured by the non-contextual representation for the relevant word (e.g., does the GloVe embedding for \emph{waiters} encode the information that \emph{waiters} is plural? masculine?). We also want to confirm that none of these tasks can be performed by non-contextual embeddings for any of the \emph{other} words of the sentence, to ensure that the information being tested for is truly contextual. We use $300$-dimensional GloVE embeddings, which prove generally adequate to encode all of the targeted word information. 

\section{Probing task results}\label{sec:probresults}

\begin{figure*}[t!]
    \includegraphics[width=16cm]{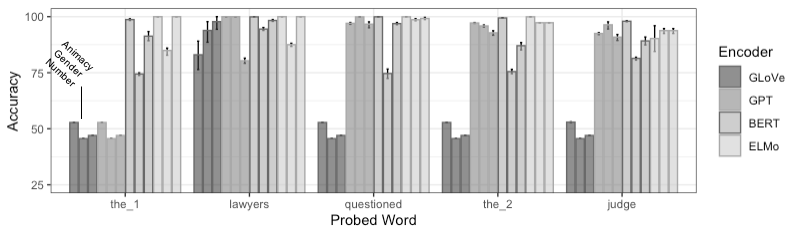}
    \centering
    \caption{Probing task results with \textbf{subject noun} as target word. Vertical ranges show 95\% confidence intervals computed with non-parametric bootstrap. Each cluster of adjacent bars of the same shade represents the three different tested information types---from left to right: number, gender, animacy}
    \label{fig:subject}
\end{figure*}

\begin{figure*}[t!]
    \includegraphics[width=16cm]{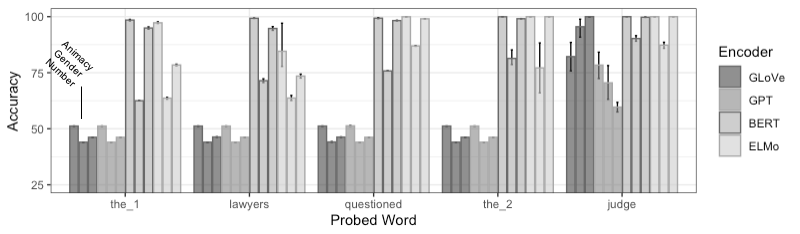}
    \centering
    \caption{Probing task results with \textbf{object noun} as target word. Vertical ranges show 95\% confidence intervals computed with non-parametric bootstrap. Each cluster of adjacent bars of the same shade represents the three different tested information types---from left to right: number, gender, animacy}
    \label{fig:object}
\end{figure*}

\begin{figure*}[t!]
    \includegraphics[width=16cm]{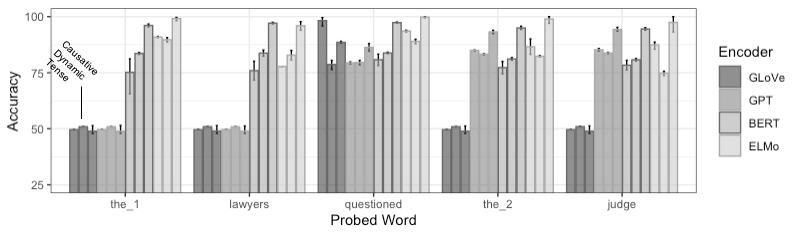}
    \centering
    \caption{Probing task results with \textbf{verb} as target word. Vertical ranges show 95\% confidence intervals computed with non-parametric bootstrap. Each cluster of adjacent bars of the same shade represents the three different tested information types---from left to right: tense, dynamic, causative}
    \label{fig:verb}
\end{figure*}

Figures~\ref{fig:subject}-\ref{fig:verb} show the results for tasks with subject noun, object noun, and verb target words, respectively (note that although the plots include tokens from an example sentence for purposes of clarity, these are results across all test sentences). Each cluster of adjacent bars of the same shade represents the three different tested information types, with left-to-right order of number-gender-animacy for noun target words, and tense-dynamic-causative for verb target words.

\paragraph{Distribution of subject noun information} Figure~\ref{fig:subject} shows the distribution of subject noun information across sentence tokens, for all three information types and for our four tested encoders. 

First, we see that our sanity-check baselines indicate that we control our datasets well: as desired, GloVe embeddings are at chance performance for every probed word apart from the target word itself---on which GloVe performance is good---and GPT is at chance to the left of the target word. This suggests that we are successfully targeting contextual information rather than spurious cues. 

Once the subject noun is encountered, GPT shows near-perfect recoverability of subject number, gender, and animacy on all of the subsequent tokens, with the strength diminishing slightly as the subject grows more distant. The exception to this strong recoverability is in animacy encoding on the subject noun itself, which is notably weaker: GPT appears to encode more information about subject animacy on the verb and object tokens than on the subject itself. Apart from this, GPT appears to distribute subject information fairly uniformly regardless of information type or probed token.

BERT and ELMo, the bidirectional contextual encoders, show more sensitivity to the interaction of information type and probed token. Both strongly encode subject number and animacy on all tokens, though BERT's encoding of animacy lags behind ELMo's in places, and both encode weaker subject information on the object noun. As for gender, BERT seemingly disregards subject gender as context information---while subject gender is near perfect recoverability on the subject noun itself, its recoverability is only around 75\% on all other BERT tokens. In contrast, while ELMo shows weak subject gender on the subject determiner and subject noun itself, it strongly encodes subject gender on the verb, object determiner, and object noun. 

\paragraph{Distribution of object noun information}
Distribution of object noun information is shown in Figure~\ref{fig:object}. Again, the validity and control of our tests is supported by chance-level performance of GloVe representations on all but the object noun, and of GPT embeddings on every token prior to the object noun. GPT shows surprisingly weak encoding of object noun information even on the object noun embedding---this pattern suggests that GPT embeddings of the object noun actually encode more information about the \emph{subject noun} several words away than about the object noun itself. 

BERT shows strong encoding of object number and animacy across tokens, but again sacrifices gender information on tokens apart from the object noun. 
ELMo also shows strong encoding of object number (with the exception of the subject noun), and of object animacy on the object noun, determiner and verb---but encodes animacy more weakly on the subject words. Unlike the case of subject gender, ELMo joins BERT in showing consistently weaker encoding of object gender.  

\paragraph{Distribution of verb information}

Distribution of information about the verb is shown in Figure~\ref{fig:verb}. Overall, encoding of verb information is weaker and somewhat more uniform across the sentence than encoding of noun information. BERT and ELMo both strongly encode the causative-inchoative alternation across all tokens of the sentence. For GPT this is also the most strongly encoded feature, and as with subject animacy, it is more strongly encoded on the later words than on the verb itself. For ELMo, the dynamic-stative property is the most weakly encoded property across the sentence (except on the subject noun). For BERT the verb's tense is the most weakly encoded, consistently lagging behind ELMo's encoding of verb tense. Among ELMo embeddings, the subject determiner shows surprisingly high performance in encoding of all verb properties. 

\paragraph{Interim summary} GPT shows uniform strong encoding of subject information and solid encoding of verb information on the target and subsequent words---but weak encoding of object information on the object noun. BERT and ELMo show more nuance in their distribution of the information types, with BERT heavily deprioritizing gender information, but strongly encoding animacy and maintaining rich number information for both nouns across all words. ELMo too deprioritizes object gender across tokens, but it shows strong encoding of subject gender after the subject noun, mostly strong encoding of animacy (apart from object animacy on subject words), and consistently rich encoding of number for both nouns. Encoding of verb features is generally weaker than noun features, with BERT weakest on tense, ELMo weakest on dynamic-stative, and all contextual models strongest on the causative-inchoative distinction. 

\section{Distance manipulation tasks}

\begin{figure*}[t!]
    \includegraphics[width=16cm]{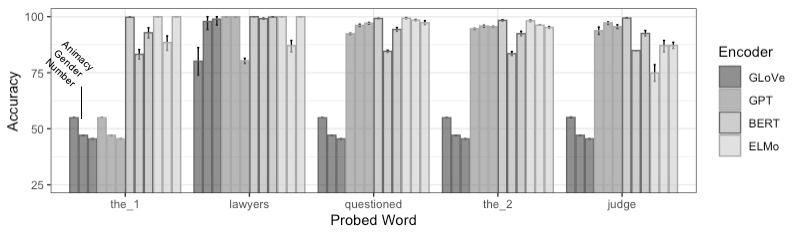}
    \centering
    \caption{Distance manipulation probing task results with \textbf{subject} as target word. Vertical ranges show 95\% confidence intervals computed with non-parametric bootstrap. Each cluster of adjacent bars of the same shade represents the three different tested information types---from left to right: number, gender, animacy}
    \label{fig:distance}
\end{figure*}

\begin{figure*}[t!]
    \includegraphics[width=16cm]{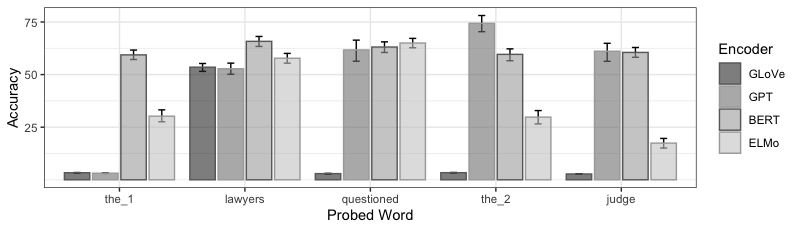}
    \centering
    \caption{Word identity task: labeling identity of \textbf{subject noun}. Vertical ranges show 95\% confidence intervals computed with non-parametric bootstrap.}
    \label{fig:wc}
\end{figure*}

\paragraph{Setup} Because our sentences follow a fixed structure for category-specific probing, it is possible that differences in encoding from word to word are an effect of linear distance rather than the syntactic/semantic relationships between the words. We perform a follow-up analysis inspired by a task in \citet{zhang2018language}, in which the authors investigate the effect of distance from the target word as a factor in how richly recurrent neural networks encode syntactic information. For all of our tasks, we introduce a manipulation to change linear distances between our target and probed words, by splicing relative clauses after the subject and adjectives before the object. For example: 
\begin{center}
        \emph{The lawyer found the judge. \\} 
        \vspace{.25cm}
        \emph{The lawyer \textbf{who was hungry} found the \textbf{angry and competent} judge.}
\end{center}
For reasons of space, we display only subject task results, in Figure \ref{fig:distance}. All results may be found in our GitHub repository linked in Footnote 1. 

\paragraph{Results} When we increase linear distances between words, the patterns remain similar to those observed in the five-word sentences. GPT still consistently encodes subject information on each of the tokens after the subject noun is encountered, with the exception of animacy encoding on the subject noun itself. BERT and ELMo still show strong encoding of subject number and animacy across tokens, with BERT dispreferring gender information across tokens and ELMo dispreferring gender only on subject determiner and noun. The main difference is that ELMo shows a marked drop in subject number information (and a bit of a drop in gender and animacy) on the object noun. 

These results suggest that the observed strong encoding of context information is not simply a function of the proximity of the words in our five-word sentences, given that the strong encoding patterns persist over the longer distances (with the slight exception of ELMo losing some encoding on the object noun). This may indicate syntactic awareness in the models, which would be consistent with the findings of, e.g., \citet{hewitt2019structural} and \citet{tenney2019you}. The results further suggest that the contextual encoders tag information as relevant to specific categories of target words in their contexts, operating flexibly across varying linear distances with different structures. 

\section{Word identity tasks}

\paragraph{Setup} We aim to show whether the encoders incorporate only more coarse-grained linguistic information in their embeddings, or if encoding is fine-grained enough to memorize the embedding patterns for specific context word identities. We use a variation of the \emph{word content} task from~\citet{conneau2018you} and~\citet{adi2016}. The goal of the original word content task is to determine whether a sentence vector representation contains a given word. We adapt this task to test the extent to which contextual embeddings can identify a neighboring word at a given position. We formulate our word identity tasks as ``What is the identity of the subject'' or ``What is the identity of the verb'', etc. As in Section~\ref{sec:probresults}, we probe each word position independently, using our fixed five-word sentences. 

For identity classification, we use a softmax \emph{k}-way classification task, similar to the word content task in \citet{conneau2018you}. The classifier for this task must choose which of the \emph{k} output words is in the target position of the sentence. In pre-testing, we found best overall performance with a $30$-way classification, for which we present the results here. Smaller and larger $k$ produce similar patterns of results, but performance overall decreases. 

\paragraph{Results} We display results for probing subject noun identity (``what is the identity of the subject noun'') in Figure \ref{fig:wc}.

This proves to be a challenging task, but we see clear trends suggesting that our encoders pick up on word identity signals. As before, GloVe embeddings are at chance on all but the target subject noun, and GPT embeddings are at chance for tokens to the left of the subject noun, satisfying our sanity checks. On the subject noun itself, encoders show comparably high recoverability of word identity, with BERT standing out as the strongest. GPT and ELMo see a slight boost in recoverability of subject identity on the verb, and GPT surprisingly shows the most subject identity information on the object determiner. BERT representations retain consistently strong subject identity encoding throughout the sentence, as do GPT embeddings starting with the subject noun itself---but ELMo encoding of subject identity drops off sharply on the determiners and object noun. This suggests that information about surrounding word identities is distributed fairly evenly across sentence tokens for BERT and GPT, but ELMo keeps word identity information fairly local to the word position itself.

Probing for the identity of the verb and of the object produces analogous patterns of results. In particular, GLoVe embeddings are at chance on all words but the target word, while GPT embeddings are at chance before the target word, and pattern similarly to BERT afterwards in the object identity task. BERT is strong throughout, while ELMo shows more effect of distance from the target word.

While identity classification performance here is far above chance, it is also well below 100\%. It is possible that performance will increase with a stronger classifier, but it is also likely that encoding of context information at the granularity of word identity is not practical or necessary for contextual embeddings, such that they more strongly encode relevant context word features rather than word identities themselves, as these results suggest. 

\section{Discussion}
The results presented here shed light on how different contextual encoders distribute information across token embeddings of a sentence. While we cannot draw strong conclusions about causal relations between model properties and the observed patterns, we can make broad connections between the two to inform future investigations. 

Overall, the deeper, transformer-based architectures of BERT and GPT do not produce dramatic differences in distribution of information relative to the shallower ELMo model---the main difference observed with ELMo's shallower recurrent architecture is a bit of a drop in information (particularly number and word identity) over longer distances, where BERT and GPT retain strong encoding. This is not necessarily surprising, given the potential of the self-attention mechanism to capture long-distance connections---it is perhaps more surprising that ELMo shows so little difference overall. These patterns suggest that deeper transformer models may not be critical for encoding and distributing these types of context information, except perhaps over substantial distances. 

BERT and ELMo, the models that use bidirectional context, generally pattern more similarly to each other than to GPT, particularly in strongly encoding number and animacy over gender, and encoding number strongest overall for nouns; GPT shows more uniformity in encoding noun information (at least from the subject noun). This pattern suggests that using bidirectional versus unidirectional context has more impact on distribution of context information than does depth or architecture type. GPT's poor encoding of object information relative to subject and verb information further suggests that the left-to-right architecture may prioritize earlier information over later information. 

As for the two bidirectional models, what BERT's particular properties seem to give it over ELMo, beyond more robustness to distance, is slightly different selectivity---dropping subject gender information earlier than ELMo does, while keeping object animacy information at a longer distance, and dropping verb tense information a bit more. Given BERT's generally stronger performance on downstream tasks, this suggests that BERT's masked language modeling setup, in tandem with its greater capacity to handle longer distances, allows for a more nuanced picture of how bidirectional context information should be distributed across tokens for optimal predictive power.

\section{Conclusion}

In this paper we have begun to tackle a key question in our understanding of the contextual embeddings on which most current state-of-the-art NLP models are founded: what is it that contextual embeddings pick up about the words in their contexts? We have introduced a novel probing approach and a suite of tasks through which we have performed systematic, fine-grained probing of contextual token embeddings for information about features of their surrounding words. We apply these tests to examine the distribution of contextual information across sentence tokens for popular contextual encoders BERT, ELMo, and GPT.

We find that each of the tested word features can be encoded in contextual embeddings for other words of the sentence, often with perfect or near-perfect recoverability. However, we see substantial variation across encoders in how robustly each information type is distributed to which tokens. Distance manipulations indicate that the observed rich contextual encoding is not an artifact of proximity between words, and probing for information about context word identities suggests a weaker encoding of identity information than of more abstract word feature information. Bidirectional context appears to impact distribution patterns more than depth or architecture, though the transformer models show more robustness to distance. Overall, these results help to clarify the patterns of distribution of context information within contextual embeddings---future work can further clarify the impact of more diverse syntactic relations between words, and of additional types of word features. We make all datasets and code available for additional testing.

\section*{Acknowledgments}
We would like to thank Itamar Francez and Sam Wiseman for helpful discussion, and anonymous reviewers for their valuable feedback. This material is based upon work supported by the National Science Foundation under Award No. 1941160.

\bibliography{acl2020.bib}
\bibliographystyle{acl_natbib}

\end{document}